\documentclass[runningheads]{llncs}
\usepackage{cite}
\usepackage{amsmath,amssymb,amsfonts}
\usepackage{algorithmic}
\usepackage{graphicx}
\usepackage{textcomp}
\usepackage{xcolor}
\usepackage{fixltx2e}
\usepackage{array}
\usepackage{gensymb}
\usepackage{dblfloatfix}

\newenvironment{conditions}
  {\par\vspace{\abovedisplayskip}\noindent\begin{tabular}{>{$}l<{$} @{${}={}$} l}}
  {\end{tabular}\par\vspace{\belowdisplayskip}}
\graphicspath{ {./images/} }
\def\BibTeX{{\rm B\kern-.05em{\sc i\kern-.025em b}\kern-.08em
    T\kern-.1667em\lower.7ex\hbox{E}\kern-.125emX}}
    
\begin{document}
\title{Application of Neuroevolution in \\ Autonomous Cars}
\author{
    Sainath G\inst{1}\orcidID{0000-0002-7401-1955} \and
    Vignesh S\inst{2}\orcidID{0000-0003-2332-9565} \and \\
    Siddarth S\inst{3}\orcidID{0000-0002-2690-381X} \and
    G Suganya\inst{4}\orcidID{0000-0001-9560-4760}
}
\authorrunning{Sainath G. et al.}
\institute{
    Vellore Institude of Technology, Chennai, India \\
    \url{chennai.vit.ac.in} \\ 
}
\maketitle              % typeset the header of the contribution

\begin{abstract}
With the onset of Electric vehicles, and them becoming more and more popular, autonomous cars are the future in the travel/driving experience. 
The barrier to reaching level 5 autonomy is the difficulty in the collection of data that incorporates good driving habits and the lack thereof. The problem with current implementations of self-driving cars is the need for massively large datasets and the need to evaluate the driving in the dataset. 
We propose a system that requires no data for its training. An evolutionary model would have the capability to optimize itself towards the fitness function. We have implemented Neuroevolution, a form of genetic algorithm, to train/evolve self-driving cars in a simulated virtual environment with the help of Unreal Engine 4, which utilizes Nvidia’s PhysX Physics Engine to portray real-world vehicle dynamics accurately. 
We were able to observe the serendipitous nature of evolution and have exploited it to reach our optimal solution. We also demonstrate the ease in generalizing attributes brought about by genetic algorithms and how they may be used as a boilerplate upon which other machine learning techniques may be used to improve the overall driving experience.
\end{abstract}


\section{Introduction}
The Society of Automobile Engineers(SAE) has coined six different levels of autonomy beginning at level 0, absence of any autonomy, to level 5, complete autonomy requiring no human intervention whatsoever.
Currently, many luxury vehicles possess level 3 autonomy in terms of cruise control and active lane control, and a handful of vehicles possess level 4 autonomy. Level 5 autonomy in cars is still under research and development. The main barrier to attain this level of autonomy is the task of collecting data and the lack thereof. Although a deep model is extremely adept at generalizing features, it can only learn what it sees. There are only so many scenarios we as humans can drive around that model can learn from. Essentially, even if it learns to navigate through a busy street, it may not be able to correct oversteer or understeer due to several factors such as poor roads, tire wear, etc. causing a loss of traction, which may not have been accounted for in the training dataset. That is why an evolutionary approach would solve these issues. What if the car could learn to drive on its own, via trial and error, over countless generations? It would have trained, evolved to overcome such edge cases and scenarios, and would know exactly what to do once it detects wheel spin or any form of loss of traction and grip.
\bigbreak
\section{Overview of Existing Systems}

\subsection{Neural Networks}
A neural network is an interconnected network of neurons, also called nodes. Each neuron has a set of output edges that activate based on the resultant value obtained from the weighted inputs it received from the previous layer.

\begin{figure}[htbp]
\centerline{\includegraphics[scale=0.25]{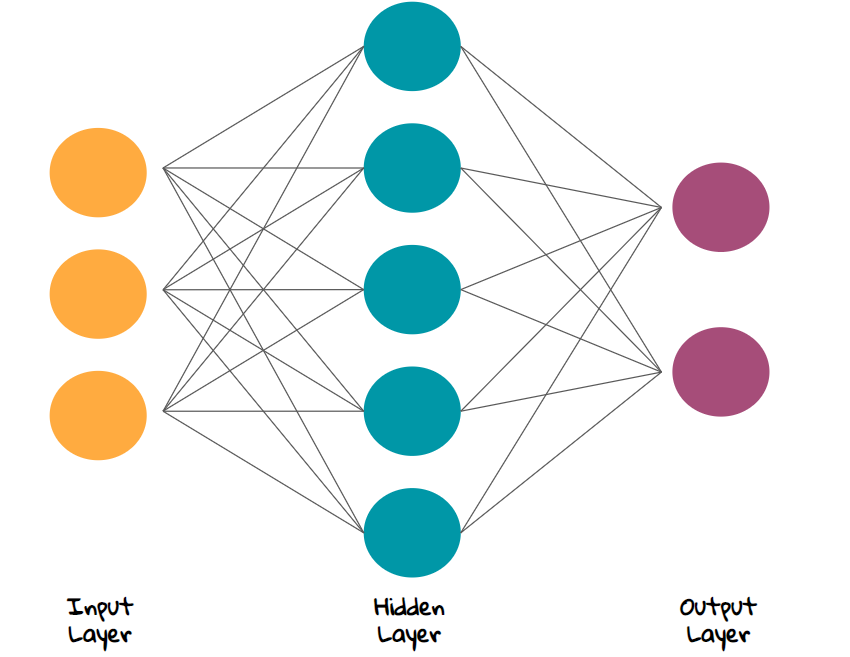}}
\caption{Topology of an Artificial Neural Network}
\label{fig}
\end{figure}

In a supervised learning approach, we would have a list of attributes or features as our inputs and a list of targets as our outputs. We would then have to use back-propagation to train our neural network to correct its weight to suit our target and increase its accuracy. Back-propagation is just a way of propagating the total loss back into the neural network to know how much of the loss each node is responsible for, and subsequently updating the weights in such a way that minimizes the loss by giving the nodes with higher error rates lower weights and nodes with lower error, greater weights.
\bigbreak
So, in a situation where it is difficult to obtain a dataset large enough to train the neural network to a certain degree of accuracy, we will face problems arriving at our optimal solution. This is especially true in the scenario of self-driving cars, where large corporations like Nvidia use 1000 hours of driving data to train their vehicle to navigate the roads.
In such scenarios, we could adopt an evolutionary technique that requires no datasets and train our model in a simulated environment.
\bigbreak
Although Deep Neural Networks utilizing convolutional layers have performed extraordinarily in several scenarios, the problem arises after the fact that it can only generalize what it is shown or taught. Since there are countless more possibilities of things that can happen on the road, which cannot be accounted for in the driving data we gather.
\bigbreak

\subsection{Reinforcement Learning}
Reinforcement Learning is another critical area of research in autonomous vehicles, where an agent learns to accomplish a task by gathering experience by itself, rather than through a supervised dataset. The basic gist of the algorithm is that an agent granted a reward when it performs an action that is desirable in the current scenario and gets punished if it does something undesirable.

\begin{figure}[htbp]
\centerline{\includegraphics[scale=0.5]{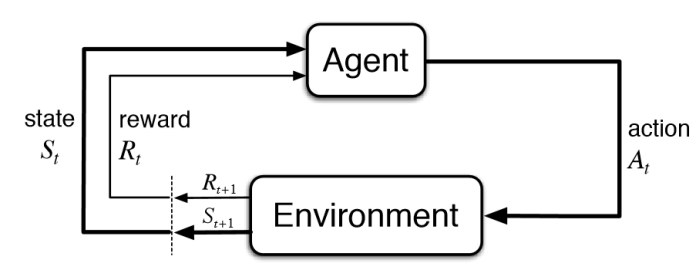}}
\caption{Basic flow of reinforcement Learning}
\label{fig}
\end{figure}

Although this form of carrot and stick approach seems to be how we, as individuals learn, the key drawback of this algorithm is that the agent has no prior experience whatsoever. We humans learn pretty quickly through this approach due to the generalization of a multitude of experiences that we have gathered from birth till date. This is not the case for the agent, and so it takes quite a while, depending on the complexity of the problem, for the agent to gather enough experiences in order for it to determine whether a certain action is good or bad.

\section{Proposed System}

\subsection{Genetic Algorithm}
Genetic Algorithms, also called Evolutionary algorithms, inspired by the process of natural selection and Darwinian evolution, mimic species evolution to arrive at a solution. Each generation has a set of species that contain specific genes, and the best are selected to populate the next generation. This process goes on until we arrive at our optimal solution.
Genetic Algorithms are currently in use to generate solutions for optimization and search problems by utilizing techniques such as selection, mutation, and crossover.

\subsection{Neuro-Evolution}
Neuro-Evolution is a Genetic Algorithm that is used to evolve artificial neural networks. In this model, each species of a generation has a brain (the neural network) that has a set of genes (weights).
In the beginning, all species of the population have random weights and hence perform random actions. It is through serendipitous discovery that a certain species gets closer to our solution. We select this species based on a fitness function and pick similarly performing species to perform crossover. After crossover, we mutate this gene and pass it on to the next generation.

So the entire genetic algorithm can be summarized to 3 key processes:
\begin{itemize}
\item Selection: We select the best species of the generation based on the fitness function.
\item Crossover: We crossover the genes of the population to converge onto our solution.
\item Mutation: We mutate the genes, in the hope of a better solution, of the selected species following crossover.
\end{itemize}

\begin{figure}[htbp]
\centerline{\includegraphics[scale=0.5]{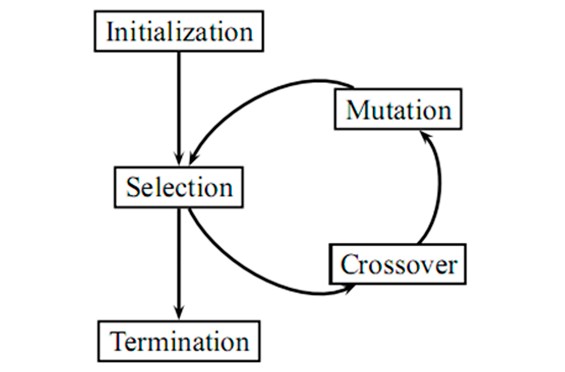}}
\caption{Genetic Algorithm Flowchart}
\label{fig}
\end{figure}

We can see that mutation and crossover seem a little opposite to one another. Mutation randomizes the weights of a certain percentage of neurons while crossover tries to converge them.
There is a trade-off here between exploration and exploitation. Exploration via mutation is exploring new gene sets out of a hope that something new can lead to promising results. While, exploitation via crossover is taking what you learned and using that information, combining the best, to inform newer decision-making processes. Typically, in genetic algorithms, the mutation rate is at 10–20\%, while the crossover rate is at 80–90\%.

\section{Implementation}
We chose to simulate Neuro-Evolution using Unreal Engine 4, which is a game engine that utilizes Nvidia’s PhysX Physics Engine to replicate real-world like vehicle dynamics, which is essential if we plan to transfer the learning that has happened in this environment.
\bigbreak
Compared to using simulators such as CARLA, which was also built on the same engine, we have a lot more freedom when we build the whole environment from the ground up, in terms of level design, vehicle physics, frame times (time dilations) and overall gives more power to the user.

\subsection{Vehicle Dynamics}
\paragraph{FWD Layout}
Since most vehicles these days in the low to mid-tier range are front-engine, front-wheel drives (FF), we chose this as our vehicle layout and for the differential, we went with a limited-slip differential (LSD) that prevents wheel spin, which is getting more and more common these days. The transmission of the vehicle is set to automatic. The suspension settings have also been altered so that it favours under-steer rather than it over-steer, as most manufacturers do these days, as it is easier to correct under-steer. Weight transfer and tyre traction are also essential aspects that dictate the vehicle’s physical handling, and are simulated accurately.
\bigbreak
\paragraph{RWD Layout}
We also wanted to observe how this approach would fare on a more difficult layout which is harder to control, which is the front engine, rear wheel drive (FR), also the typical sports car layout, as they more prone to over-steering and sliding through corners without proper throttle control and adequate counter-steering. The suspension of this layout has also been altered so that it favours over-steering behaviour rather than under-steer.
\bigbreak

\subsection{Neural Net}
Each vehicle we simulate has a brain that controls the values for the throttle pedal, the brake pedal, and the steering angles directly. This brain is our deep neural network which outputs a value from $-1\;to\;1$ for all the above inputs of the vehicle.
\bigbreak
The inputs to the neural network are the distances (normalized to $0\;to\;1$) we obtained by tracing a point cloud around the vehicle. We also feed in the current speed of the vehicle (normalized to its maximum speed) and the angle between the velocity of the vehicle and its forward vector, which provides the neural net information about in which direction the car is sliding towards, if or when it does.
\bigbreak
\subsection{Genetic Algorithm}
The genetic algorithm in this scenario is a higher-level entity that oversees the processes responsible for selection, crossover and mutation. It controls the mutation and crossover rates and is responsible for spawning and tracking all the features of the entire vehicle species for each generation of the population.
\bigbreak

\begin{figure}[htbp]
\centerline{\includegraphics[scale=0.7]{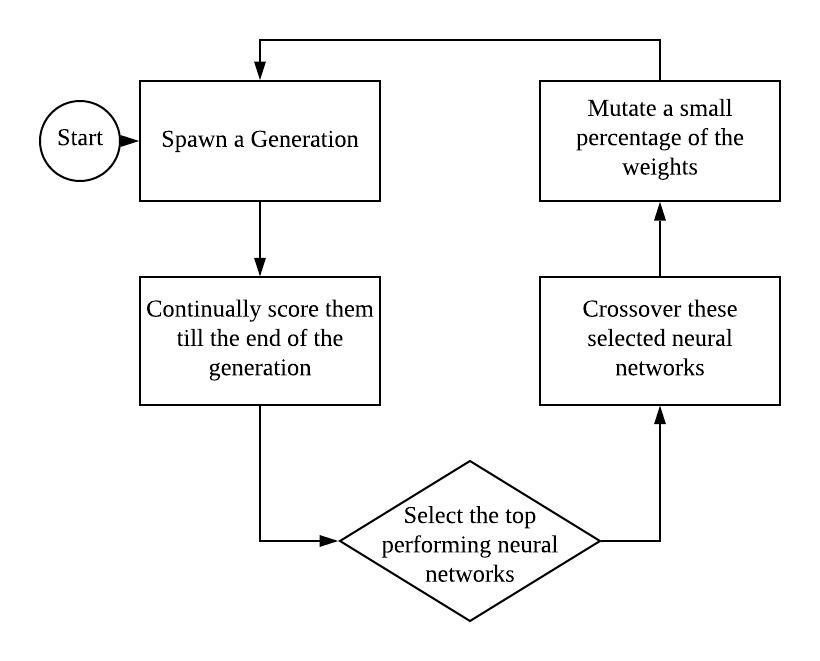}}
\caption{Simple Pipeline of Neuroevolution}
\label{fig}
\end{figure}

Initially, in the first generation, all the weights of all the neural networks are initialized to random values, and it is through serendipitous discovery aided by the fitness function that we converge on to a solution through selection and crossover and also search for a better solution through mutation. 
\bigbreak
\subsection{Working}
In the first generations, all the weights of all the neural networks in the vehicles are initialized randomly, so they have no clue what to do when they are spawned and hence move randomly. To remove poorly performing agents, which is a crucial part of Darwinian evolution, “survival of the fittest”, we de-spawn vehicles that crash into obstacles and guard rails or those that do not reach a certain threshold score within a predetermined period of time.
\bigbreak
The working of the whole model follows the following pipeline:

\begin{enumerate}
    \item The Genetic Algorithm agent spawns a population
    \item The top-performing vehicles are selected via the fitness function
    \item They are then crossover-ed by the weighted average of their weights in the neural network which is obtained by multiplying them with their relative fitness with respect to the population
    \item Once crossover is done, we mutate a small percentage of the weights by setting them to random values.
    \item We now spawn the next generation of vehicles
\end{enumerate}
This process is repeated for several generations till we obtain satisfactory results.
\bigbreak

\begin{figure*}[htbp]
\centerline{\includegraphics[scale=.6]{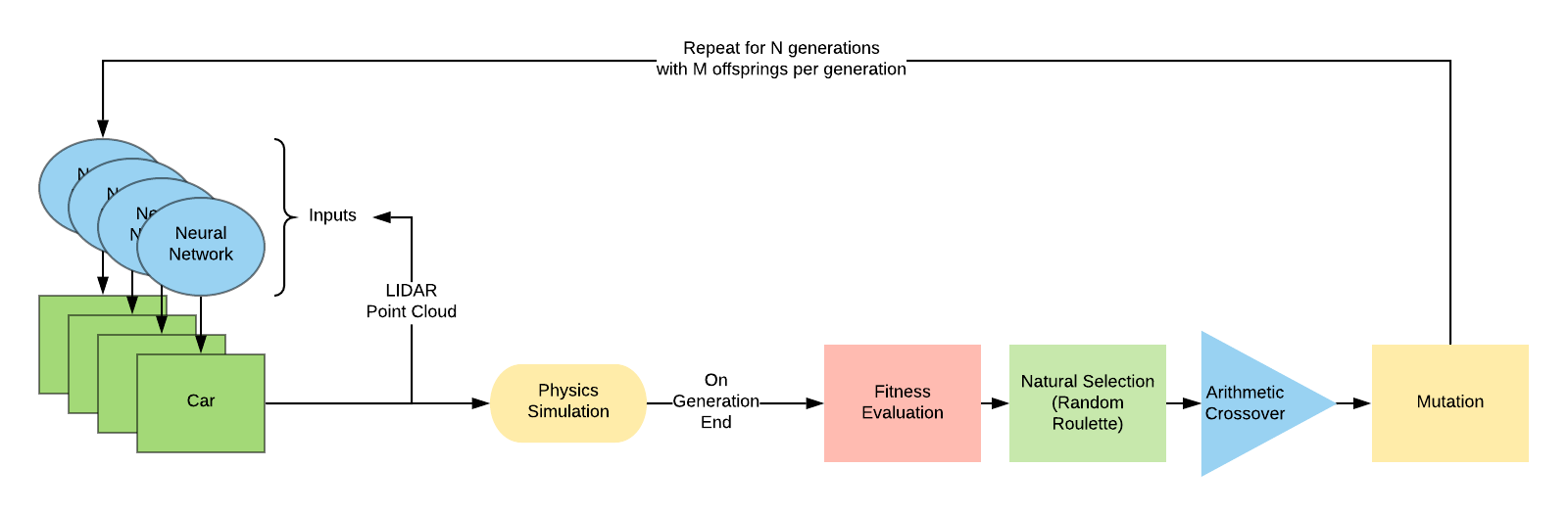}}
\caption{Neuroevolution Architecture}
\label{fig}
\end{figure*}

\paragraph{Selection}
For each vehicle, the distance travelled (in the direction of the course) each frame, which we call the score of the neural net and is calculated as:
\begin{equation}
\Delta d = v \times \Delta t 
\end{equation}
\begin{conditions}
\Delta d    &  distance travelled that frame \\
v           &  instantaneous speed \\
\Delta t    &  frame time
\end{conditions}
In order to prevent over-correcting behaviour and so that it doesn't game the fitness score, we increment its score only when the angle between the velocity vector and the car's forward vector is less than a threshold value which we set as $10\degree$.
\bigbreak
From this, we calculate the net score each frame, which is the total distance travelled (until it de-spawns) and is calculated as:
\begin{equation}
score = \sum \Delta d
\end{equation}

At the end of each generation, the relative fitness of each neural network is calculated as:
\begin{equation}
fitness_{i} = \frac{score_{i}}{\sum_{j=1}^{p} score_{j}}
\end{equation}
where:
\begin{conditions}
fitness_i   &  relative fitness of the current neural network \\
score_i     &  total distance travelled by the vehicle \\   
p           &  total population of the generation
\end{conditions}

Now for spawning the next generation of vehicles, we pick the top $n$ vehicles with the greatest fitness ($n$ could be selected arbitrarily, we chose it to be $1/10^{th}$ of the population, $p$).
\bigbreak
\paragraph{Crossover}
We now perform an arithmetic crossover of these $n$ species by weighted addition of their weights with respect to their fitness.
For each connection in the neural network, the weight of the connection after crossover is calculated as:
\begin{equation}
new\;w = \sum_{i=1}^{n} (w_{i} \times fitness_{i})
\end{equation}
where:
\begin{conditions}
fitness_i   &  relative fitness of the current neural network \\
w_i         &  weight of a certain connection in the neural net \\
n           &  the top selected species of the generation
\end{conditions}

\paragraph{Mutation}
Once we perform crossover for about 80\% of the weights in the neural network, we then move on to mutation which is typically done to about 20\% of the weights by using a random function on the weights.

\section{Results}

Within the simulated environment, we have observed that the population over several generations have evolved to not crash into obstacles, and also through sheer randomness have decided to stick to one side of a lane in certain simulations. They also perform advanced traffic management techniques such as zipper merges. A zipper merge is when a car continues to stay on the lane even after a blockade is located. They merge into the free lane only once they are close to the blockade in order to prevent traffic congestions that would occur if everyone stopped using that lane entirely.

\begin{figure}[htbp]
\centerline{\includegraphics[scale=0.3]{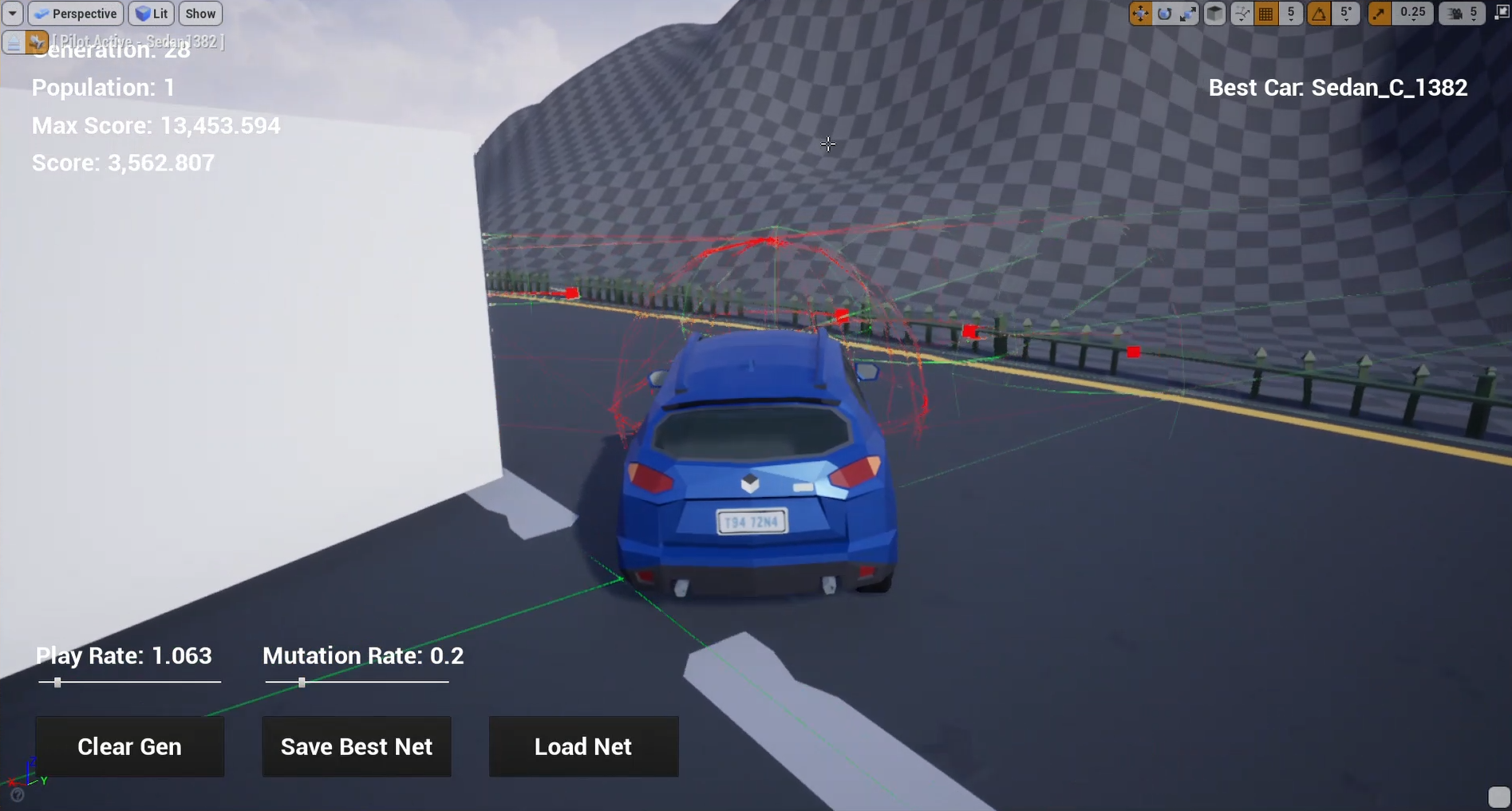}}
\caption{The neural net switches lane only at the very verge of colliding on to the obstacle(white wall to left)}
\label{fig}
\end{figure}

\begin{figure}[htbp]
\centerline{\includegraphics[scale=0.3]{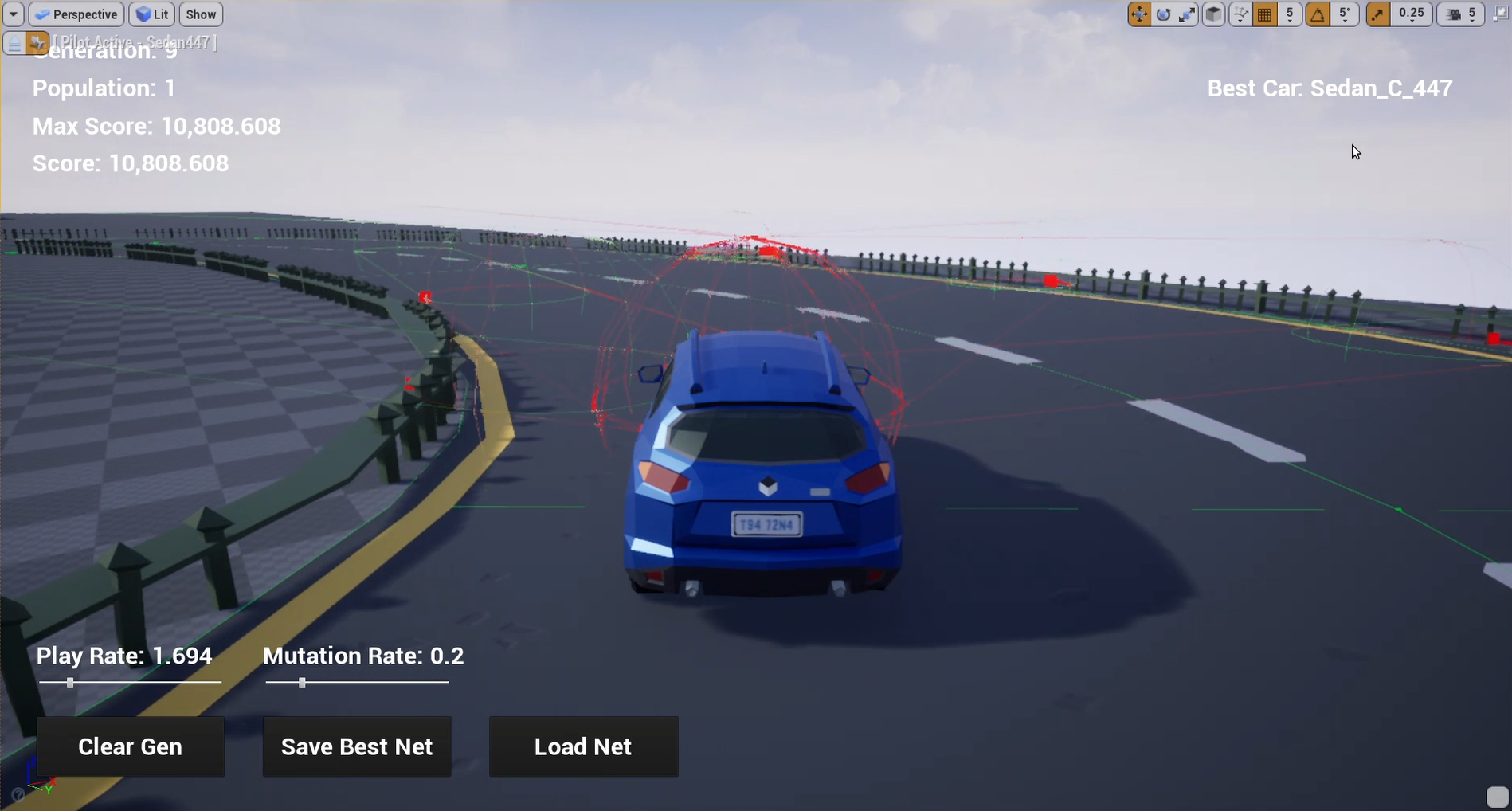}}
\caption{The neural net has decided to stick to the left lane on the road through sheer randomness which can nurtured by altering the fitness function}
\label{fig}
\end{figure}

\begin{figure}[htbp]
\centerline{\includegraphics[scale=0.45]{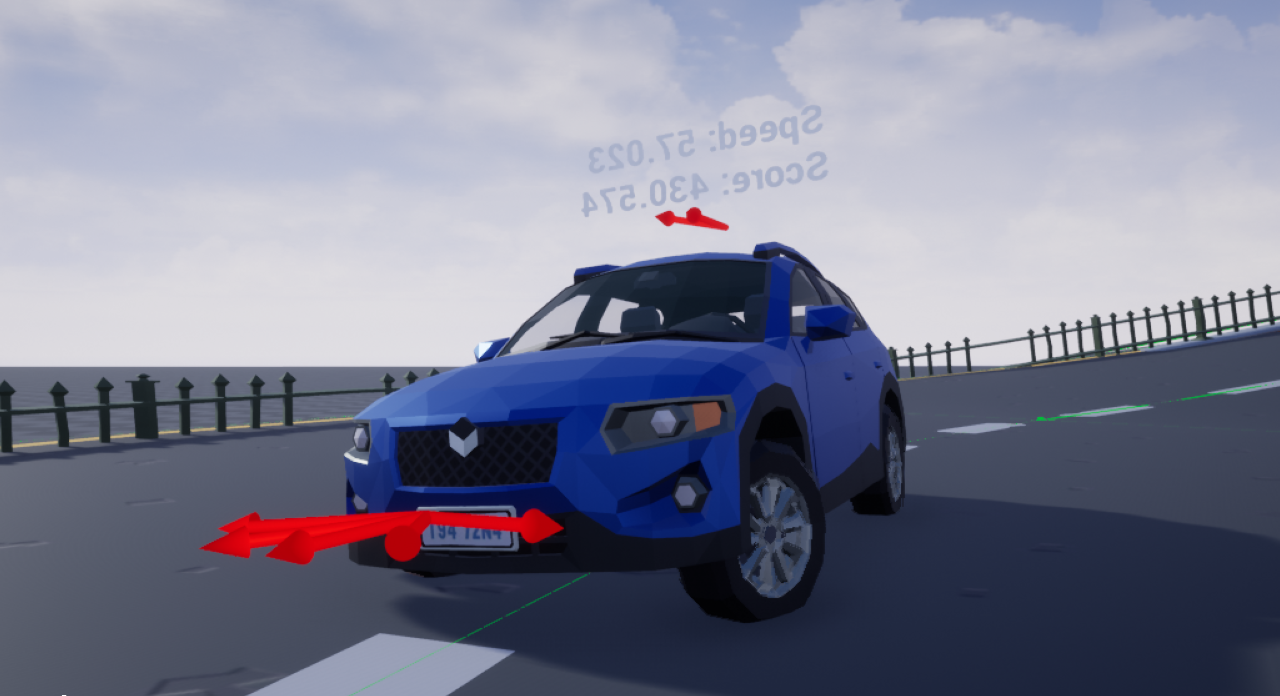}}
\caption{The neural net has learned how to counter-steer and control the car on the onset of over-steer}
\label{fig}
\end{figure}

\begin{table}[htbp]
\caption{Generations taken by the neural nets to evolve enough to navigate the entire course without crashing}
\begin{center}
\begin{tabular}{|l|l|l|l|l|} 
    \hline
    \textbf{Layout} & \textbf{Crossover Rate} & \textbf{Mutation Rate} & \textbf{Generation} & \textbf{Population} 
    \\ [1ex]
    \hline
    FR & 80\% & 20\% & 97 & 4850 \\
    \hline
    FR & 80\% & 10\% & 225 & 11250 \\
    \hline
    FR & 90\% & 20\% & 145 & 7250 \\
    \hline
    FR & 90\% & 10\% & 171 & 8550 \\
    \hline
    FF & 80\% & 20\% & 24 & 1200 \\
    \hline
    FF & 80\% & 10\% & 26 & 1300 \\
    \hline
    FF & 90\% & 20\% & 38 & 1900 \\
    \hline
    FF & 90\% & 10\% & 12 & 600 \\
    \hline
\end{tabular}
\label{table:1}
\end{center}
\end{table}

\begin{figure}[htbp!]
\centerline{\includegraphics[scale=0.6]{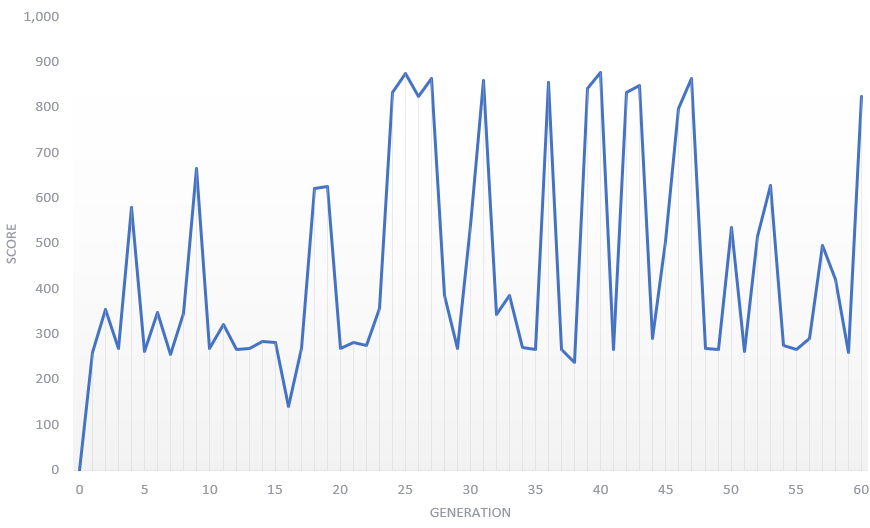}}
\caption{Once one the neural nets hits a peak, it is able to constantly replicate the peaks which is an indication of evolution}
\label{fig}
\end{figure}

A few things we observed over the course of running several simulations, iterating and variating different parameters, is that when the fitness function is simple, it generalizes the course pretty quickly and is able to navigate it well over very few generations. But, when we alter it so that it favours a certain style of cornering or maintaining a certain amount of speed, it takes drastically more generations for it achieve this sort of specialization. This interpretation is backed by the discrepancies seen in the number of generations it took for the Front engine, Rear wheel drive (FR) layout compared to the Front engine, Front wheel drive(FF) layout.
\bigbreak
\bigbreak
\section{Conclusion and Future Work}
Based on the results we've observed above, we can come to the conclusion that genetic algorithms such as NeuroEvolution can speed up the initial phase of generalizing features several fold compared to traditional techniques such as back-propagation which place the prerequisite of procuring a massive dataset for training so as to not over-fit the solution. But once the network attains the basic cognitive abilities for driving, it can be further improved upon through reinforcement learning techniques such as Deep Q-learning, since it has already gathered a plethora of experiences over several generations, which is one of the key barriers slowing down reinforcement, since now we can quickly jump to the phase where the focus is more on obtaining as many rewards as possible rather than the initial phase of gathering experience where the agent primarily tries to just not get punished for its actions.


\begin{thebibliography}{00}
\bibitem{}Bimbraw, Keshav. (2015). Autonomous Cars: Past, Present and Future - A Review of the Developments in the Last Century, the Present Scenario and the Expected Future of Autonomous Vehicle Technology. ICINCO 2015 - 12th International Conference on Informatics in Control, Automation and Robotics, Proceedings. 1. 191-198. 10.5220/0005540501910198. 
\bibitem{} Such, F.P., Madhavan, V., Conti, E., Lehman, J., Stanley, K.O., \& Clune, J. (2017). Deep Neuroevolution: Genetic Algorithms Are a Competitive Alternative for Training Deep Neural Networks for Reinforcement Learning. ArXiv, abs/1712.06567.
\bibitem{} Mardle, Simon \& Pascoe, Sean \& Tamiz, Mehrdad. (2000). An investigation of genetic algorithms for the optimization of multi‐objective fisheries bioeconomic models. International Transactions in Operational Research. 7. 33 - 49. 10.1111/j.1475-3995.2000.tb00183.x. 
\bibitem{b} Stanley, Kenneth \& Clune, Jeff \& Lehman, Joel \& Miikkulainen, Risto. (2019). Designing neural networks through neuroevolution. Nature Machine Intelligence. 1. 10.1038/s42256-018-0006-z. 
\bibitem{} Long Zhao and Zemin Liu, "A genetic algorithm for reinforcement learning," Proceedings of International Conference on Neural Networks (ICNN'96), Washington, DC, USA, 1996, pp. 1056-1060 vol.2, doi: 10.1109/ICNN.1996.549044.
\bibitem{} Iglesias Rodriguez, Roberto \& Rodríguez, Miguel \& Regueiro, Carlos \& Correa, José \& Barro, S.. (2006). Combining reinforcement learning and genetic algorithms to learn behaviours in mobile robotics.. 188-195. 
\bibitem{} Zong, Xiaopeng \& Xu, Guoyan \& Yu, Guizhen \& Su, Hongjie \& Hu, Chaowei. (2017). Obstacle Avoidance for Self-Driving Vehicle with Reinforcement Learning. SAE International Journal of Passenger Cars - Electronic and Electrical Systems. 11. 10.4271/2017-01-1960. 
\bibitem{} Sáez, Yago \& Perez Liebana, Diego \& Sanjuán, Oscar \& Isasi, Pedro. (2008). Driving Cars by Means of Genetic Algorithms.. 1101-1110. 
\bibitem{} AbuZekry, Ahmed. (2019). Comparative Study of NeuroEvolution Algorithms in Reinforcement Learning for Self-Driving Cars. European Journal of Engineering Science and Technology. 10.33422/EJEST.2019.09.38. 
\bibitem{b9} Floreano, Dario \& Mondada, Francesco. (1999). Automatic Creation of an Autonomous Agent: Genetic Evolution of a Neural-Network Driven Robot. 
\bibitem{} Chang, Simyung \& Yang, John \& Choi, Jaeseok \& Kwak, Nojun. (2018). Genetic-Gated Networks for Deep Reinforcement. 
\bibitem{} Sehgal, A., La, H.M., Louis, S.J., \& Nguyen, H. (2019). Deep Reinforcement Learning Using Genetic Algorithm for Parameter Optimization. 2019 Third IEEE International Conference on Robotic Computing (IRC), 596-601.
\bibitem{} Ahmad, Tanwir \& Ashraf, Adnan \& Truscan, Dragos \& Porres, Ivan. (2019). Exploratory Performance Testing Using Reinforcement Learning. 10.1109/SEAA.2019.00032. 
\bibitem{} Kardell, Simon and Mattias Kuosku. “Autonomous vehicle control via deep reinforcement learning.” (2017).
\bibitem{} Stanley, Kenneth \& Miikkulainen, Risto. (2002). Efficient Reinforcement Learning through Evolving Neural Network Topologies. 
\end{thebibliography}
\end{document}